\documentclass[a4paper,conference]{IEEEtran}
\pdfoutput=1

\usepackage{amsmath,amssymb,amsfonts}
\usepackage{graphicx}
\usepackage{xcolor}
\usepackage{tikz}
\def\checkmark{\tikz\fill[scale=0.4](0,.35) -- (.25,0) -- (1,.7) -- (.25,.15) -- cycle;}

\ifCLASSINFOpdf
\else
\fi
\usepackage{algorithmic}

\begin{document}
%
\title{THORN: Temporal Human-Object Relation Network for Action Recognition}

\author{\IEEEauthorblockN{Mohammed Guermal, Rui Dai, and François Brémond}
\IEEEauthorblockA{
Inria, Université Côte d'Azur, 2004 Route des Lucioles, 06902 Valbonne \\
\texttt{\{mohammed.guermal, rui.dai, francois.bremond\}@inria.fr}}}

%


\maketitle

\begin{abstract}
Most action recognition models treat human activities as unitary events. However, human activities often follow a certain hierarchy. In fact, many 
human activities are compositional. Also, these 
actions are mostly human-object interactions. In this paper we propose to recognize human action  by leveraging the set of interactions that define an action. In this work, we present an end-to-end network: THORN, that can leverage important human-object and object-object interactions to predict actions. This model is built on top of a 3D backbone network. The key components of our model are: 1) An object representation filter for modeling object. 2) An object relation reasoning module to capture object relations. 3) A classification layer to predict the action labels. To show the robustness of THORN, we evaluate it on EPIC-Kitchen55 and EGTEA Gaze+, two of the largest and most challenging first-person and human-object interaction datasets. THORN achieves state-of-the-art performance on both datasets.

\end{abstract}

%
\IEEEpeerreviewmaketitle

\section{Introduction}

Human activity recognition in video is a fundamental problem in computer vision, due to its large field of applications, such as human-computer interaction~\cite{human-computer} or video surveillance~\cite{video-surveillance}. Machine learning and computer vision models have achieved interesting results in this field. Unfortunately, most of the State-of-the-art methods focus on simple 
activities such as \textit{walking} or \textit{drinking}, while the recognition of longe-term, complex, and composite activities such as \textit{assembling furniture} or \textit{food preparation} has been rarely addressed. These methods make use of end-to-end models that produce a video level label, and do not explicitly decompose the action into a hierarchical set of sub-actions or interactions. Moreover, neuroscience \cite{barker1951one,barker1955midwest} has shown that the human perception of action is actually based on decomposing an action into different groups of interactions which enables him/her to understand other human behaviors. In this paper we decide to visit this composite actions, that we refer to as actions of Human-Object Interaction (HOI). Not only that we also focus on first-person view HOI action recognition. 

first-person action recognition also comes with its challenges, one of which is the narrow field of view that makes actions sometimes happen outside the video viewing range. Also, the huge ego-motion caused by the sharp movements of the camera can make it harder to recognize actions. Finally, in ego-vision, the field of view usually covers the human hands and an ensemble of objects. In this case, actions are generally involving interactions between the human and objects. Hence the challenge is also to recognize which of these objects are relevant to the action and which are distractors.

\begin{figure}[!t]
\centering
\includegraphics[width=\linewidth]{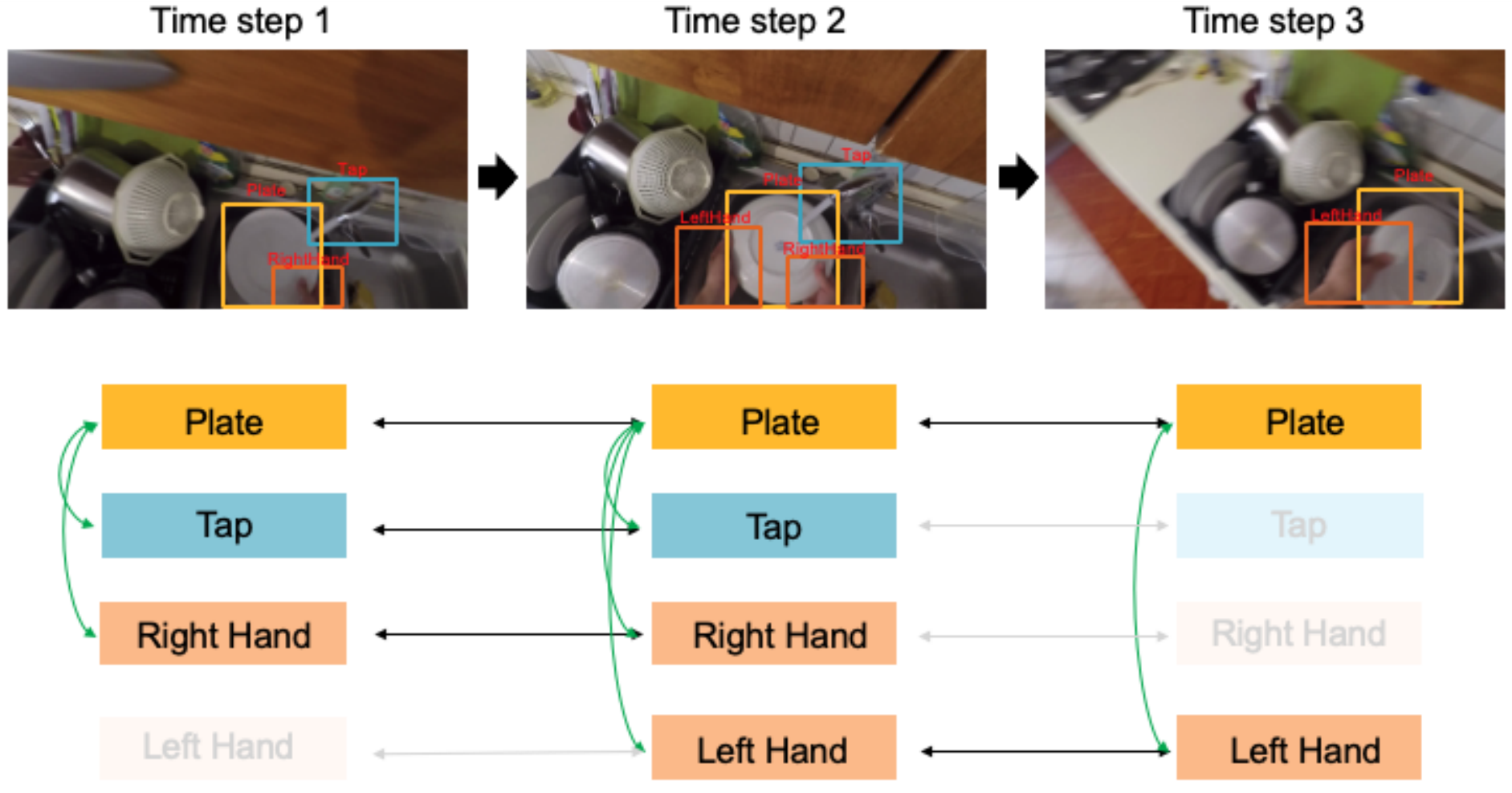}
\caption{An example of the Human-Object Interactions of \textit{wash plate} in an first-view video. Green arrows represent interactions at the same time step (i.e., spatial relation) while black arrows represent interactions across time. In practice, the model captures all the objects detected. For simplicity, here we highlight only the relevant objects to \textit{wash plate}. }
\label{fig_front}
\end{figure}

A HOI action can be seen as combination of verbs and nouns, for instance the action \textit{cutting bread with knife} is the combination of the verb \textit{cut} and the nouns \textit{knife} and  \textit{bread}. Hence recognizing an action of HOI, is a class of visual relationship detection, where the task is to not only recognize the objects (the noun), but also to infer the relation and motion (the verb) between different objects and the human. Fig.~\ref{fig_front} represents an example of an object-based action: \textit{wash plate}. Such action requires highlighting objects like the \textit{hand}, the \textit{plate} and the \textit{tap} while giving less attention to other objects that are not important to the action.

Previous works such as two-stream CNNs \cite{two-stream-1},\cite{two-stream-2},\cite{two-stream-3} or 3D CNNs \cite{3D-cnn-1} \cite{3D-cnn-2} \cite{3D-cnn-3} have achieved very good results on third view and video level label datasets~\cite{vd1}~\cite{vd2}~\cite{vd3}~\cite{vd4}.
However, when it comes to HOI actions they still lack in performance. That is due mainly to the fact that CNNs capture shareable local features in the image/videos, and they can not handle complex or fine-grained actions. Another major challenge is the fact that such activities  can often be performed in a wide variety, making it harder for CNNs to learn significant patterns.

Thus, our intuition is to build a model that can, extract detailed and object specific semantics in the videos, as well as explore the cross-object relation at different time-steps. By doing so we can firstly, improve object recognition in actions of HOI (the noun). Moreover, we can refine the motion recognition (the verb) by having a clearer idea about the interaction of these objects and their roles in the action. Finally, by encoding the scenes into a graph of objects interactions, we make it easier to learn patterns for actions even if they have many variations, since the interactions are usually the same.

To step-up to the aforementioned challenges, we propose a new module built on top of 3D-CNNs, this module is divided into two sub parts. Firstly, we design an \textbf{Object Representation filter}. This first sub-module acts as a filter that retrieves specific and object-related semantics from the overall and mixed representation (extracted from the 3D-CNN). Secondly, we add an \textbf{Object Relation Reasoning} module that uses the detailed representations to explore cross-objects relations (interactions). Finally, we obtain an object-centric model that can predict actions of HOI by exploring human-object and object-object interactions.\\


To summarize, our main contributions are:
\newline
1. A model that can find and extract detailed semantics of specific objects;
\newline
2- A graph-based module capable of exploring interactions between different objects.\\

\section{Related work}


Human-object interaction action recognition became the focus of many research subjects lately, especially with the development of important datasets such as~\cite{Epic-kitchen,ego-charades,fathi2012learning}. Several approaches have been proposed to tackle this problematic. In the following, we review some of these approaches.

\subsection{3D-CNNs}
3D-CNNs methods focus on getting the overall appearance of the videos  without considering the objects interactions. Since these methods cannot capture specific or detailed semantics, they are still limited in case of actions of HOI. Making this architectures more adequate to video level labels. We cite as an example I3D \cite{3D-cnn-2}. Although it achieves good results on many  action recognition datasets, its performance is still poor on actions of HOI. To improve the performances on these 3D-CNNs, Long Features Bank \cite{LFB} for instance, tries to capture HOI actions by extracting and fusing features from local clips as well as globally from the whole video. 
This method uses object detection and ROI-Align to capture detected object features. And though they successfully capture richer features and more temporal information, they fail to do any object interaction modeling. Hence, they cannot improve much on  HOI actions. In the same direction, Temporal Binding Networks (TBN) \cite{TBN} proposes to capture local clip features from different clips and fuse them for later prediction. In addition to that, TBN uses multi-modalities as they capture audio-visual features using audio, RGB, and optical flow. However, we believe that this multi-modality will not always bring much information about the objects. sounds can be very noisy and very similar which can confuse the prediction. Moreover, fusing multi-modalities can be hard and requires lot of efforts the may not lead to significant improvements. Finally, other works such as \cite{Anet} use also multi-modality reasoning. However, we argue that HOI actions recognition requires more focus on objects and their interactions. 

\subsection{Graph's Convolutions}

Recently, graphs have also been considered a way for solving action recognition~\cite{video-as-graph,agcn,Gskel,CTRN}. 
\par
As for human-object interaction, videos as a space-time region graph~\cite{video-as-graph} propose to model the interaction between objects and humans in two steps as they build two different graphs. This allows to correlate objects across space-time. Similarly, in \cite{Stacked_STGCN},  the authors construct the nodes of the graph with consideration to the node class. For instance, the node for the scene is computed using the aforementioned I3D. While for objects, they use the Faster-RCNN network~\cite{fastcnn} trained on MS COCO.
All these methods mentioned above try to define their nodes by using ROI-Align.
However, this is not optimal as,in most cases, multiple objects are present at the scene and some of them are too close to each other. In this case, the projected coordinates of different objects tend to be in the same set of pixels. Therefore, extracting an object's specific feature from a feature map with low resolution becomes difficult. Not only that these methods rely on pre-trained object detetctors, hence they can not leverage only objects relevant to the action. Whereas in our work, we learn to filter only relevant objects and learn specific representation to different object-classes in an end-to-end way.

\par
\par
In the domain of semantic modelling, Class Temporal Relational Network (CTRN)~\cite{CTRN} is proposed for the action detection tasks. However, CTRN is a two steps method, which is built on top of pre-extracted flattened 1-dimensional features. The dissociation between the visual encoder and temporal module makes the model overlook the appearance and spatial information in the video, while such information is critical to the HOI action recognition. 
%
%
In this work, we propose a one-step method THORN for HOI action recognition. Different from CTRN, our method leverages the object detector to extract the object semantics directly from the spatio-temporal features. After that, graph reasoning is applied to refine the object representation and to jointly model inter-object relations. 
This design allows the model to capture the latent relations among the objects in the videos, which results in higher accuracy in HOI action recognition. 
%
\begin{figure*}[!t]
\centering
\includegraphics[width=.7\linewidth,height=4cm]{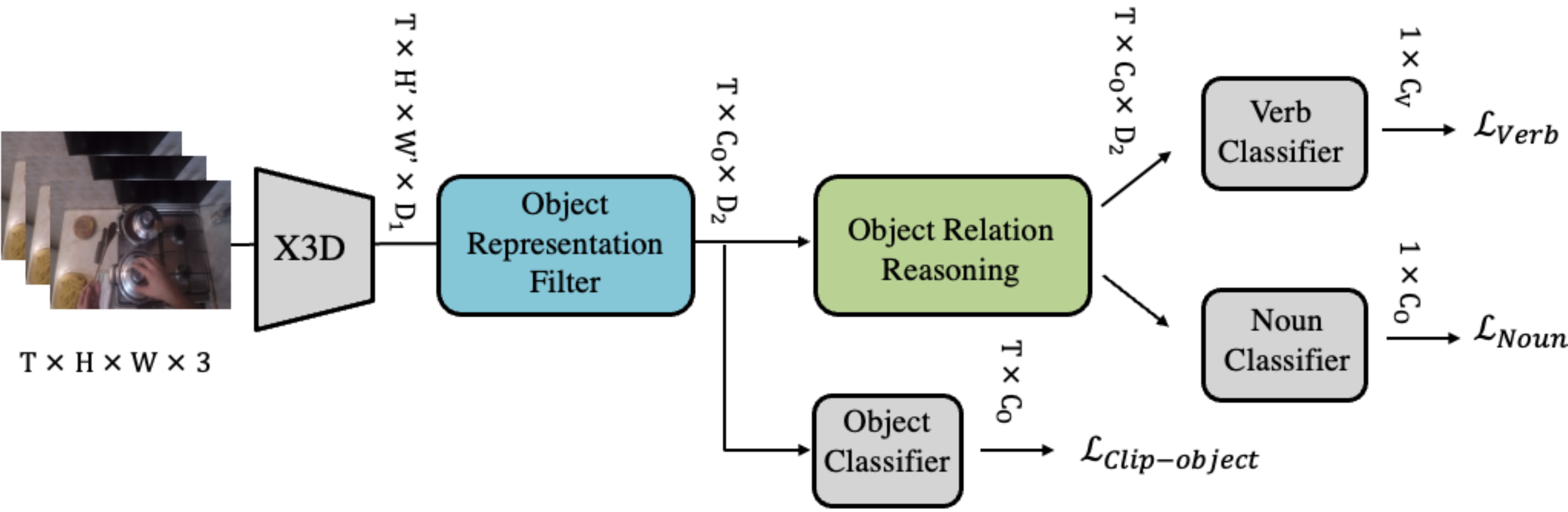}
\caption{
THORN architecture contains three main components:
(1) a \textbf{Visual encoder} (i.e., X3D) encodes the input RGB clip into a primary spatio-temporal representation. (2) The obtained representation is fed to the \textbf{Object Representation Filter}, which maps the previous representation into object-class representation. To ensure a discriminative object representation, an object classifier is added on top of the object-class representation. This classifier is trained with the pseudo-object ground truth provided by an object detector. (3) The object-class representation is also sent to the \textbf{Object Relation Reasoning} module to model the temporal-object relation in a dissociated manner.
Finally, two classifiers are used to predict the verbs and nouns relevant to the action.}
\label{fig_framework}
\end{figure*}

\section{Proposed Method}
In this section, we detail each sub-part of the proposed model, THORN. The main components in this model are: a \textbf{3D Visual Encoder} which encodes the video into a spatio-temporal embedding. Then, the previously extracted embeddings are passed to the \textbf{Object Representation Filter} (ORF). This filter extracts class-specific features. Finally, the \textbf{Object Relation Reasoning} module computes the relation between the different objects to predict the action. Fig.~\ref{fig_framework} provides an overview of the model.

\subsection{Visual Encoder}
We start by using a visual encoder to extract an embedding that serves as a full understanding of the scene, and carries the global information of the input frames. We choose X3D \cite{X3D} as our visual encoder. 
X3D has many advantages as it does not do any temporal pooling and keeps the full temporal information, providing richer temporal information. Moreover, X3D is a lighter model compared to other architectures such as I3D \cite{3D-cnn-2}.
The input to the 3D encoder is a set of video-clip frames. The output is a spatio-temporal representation $F$ of shape $(T \times H' \times W' \times {D_1})$, where:
$H'$ = $W'$ = 7, $D_1$ = 432, while $T$ is the same as the input.
\newline
This embedding carries both spatial and temporal information. The spatial information is important, as it provides object related information, such as its appearance, shape and position (e.g. drawers usually appear at the bottom of the image). That is why instead of using the X3D final output of shape $(T   \times 2048)$ to construct our nodes, we use a finer spatial representation of shape $(T   \times 7 \times 7 \times 432)$, making nodes of our graph contain more and finer information about the objects. We provide more details on this in the ablation study, by comparing both settings.
Finally, as X3D is a light-weighted model it is easier to train the \textit{Visual Encoder} jointly with the following modules.
\subsection{Object Representation Filter}
\begin{figure}[!t]
\includegraphics[width=\linewidth]{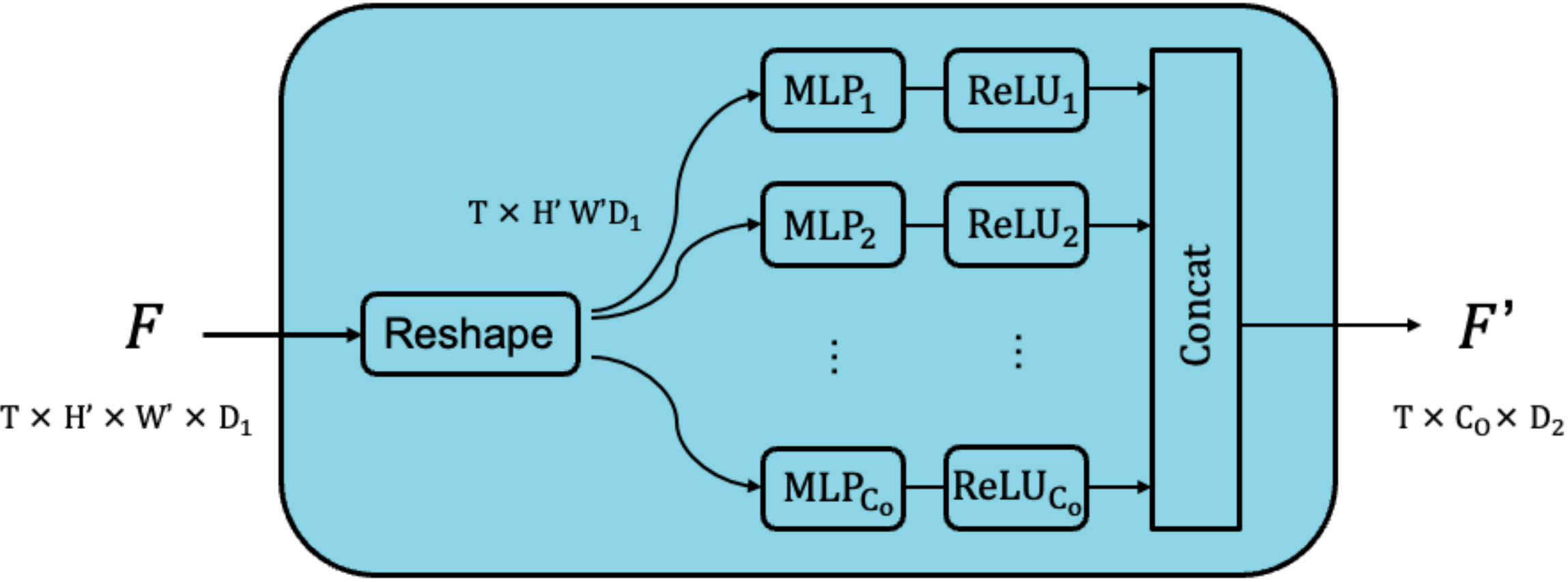}
\caption{Representation of our Object Representation Filter (ORF). The input is the feature map from the 3D encoder reshaped to $T \times H'W'D$ and the duplicated $C_{o}$ times, where $C_{o}$ is the number of classes. Finally, we have a representation specific to each object class. }
\label{fig_orf}
\end{figure}

Our main objective through this work is to have object-based reasoning. Hence the first step is to obtain object in scene representations. Therefore, we developed the \textit{Object Representation Filter} module, capable of extracting semantic representation specific to each object class from the previous overall representation. 
This module serves as a filter to obtain the object-specific representation from the output of visual encoder. 
In practice, firstly, we reshape the representation $F$ from the visual encoder to shape $(T \times H'W'{D_1})$. After that, we duplicate the reshaped features $F'$ for $C_o$ times, where $C_o$ indicates the number of object classes in the dataset. 
For each class, we use a channel-mixer MLP (i.e., linear transformation layer), 
followed by non-linear activation and dropout. In Fig.~\ref{fig_orf}, we show an overview of the ORF module. We argue that each MLP layer learns to filter features specific to a certain object class.
The equations in this module can be formulated as:
\vspace{0.005cm}
\begin{equation} \label{eq:1}
F'_{i}=ReLU(MLP(F))
\end{equation}
\begin{equation} \label{eq:2}
F'= DropOut([F'_{1},F'_{2},F'_{3}....,F'_{C_o}])
\end{equation}
With $F'$ $\in$ \hspace{0.1 cm}  $\mathbb{R}^{T \times C_o \times D_2}$. Where $D_2$ is smaller than $D_1$  to shallow the channel size. Finally, we add another MLP layer on top of $F'$ that would represent the object classifier in Fig.~\ref{fig_framework}.
\begin{equation} \label{eq:3}
F'' = ReLU(MLP(F'))
\end{equation}

Here $F''$ $\in$ \hspace{0.1 cm} $\mathbb{R}^{T \times C_o \times 1}$. 
To ensure the object-specific representation, we add a frame-level object classifier on $F''$. As the frame-level object label is not provided by the dataset, the object classifier is trained with the pseudo label provided by an object detector (i.e. Fast-RCNN~\cite{fastcnn}). In the video, multiple objects can appear in a frame, thus, we train the object classifier with binary cross-entropy loss: $\mathcal{L}_{clip-objects}$. Finally the ORF module outputs a representation for each object-class. However, we still need to correlate and refine these object representations to explore their interactions and model the actions. To do so, we introduce the next module of our pipeline in the next section.

\subsection{Object Relation Reasoning Module}
To correlate between the aforementioned representations in the previous section, we introduce the \textit{Object Relation Reasoning Module}.\\
In order to extract the relations 
between the filtered object classes, we propose to make use of graph convolutions. In the previous section, we transform the clip representation into a class-specific representation. Then, we map it to a graph-like structure, where each vertex of the graph represents an object class at a time step; the vertex  would be the previously extracted embedding of a certain class. In total, the graph consists of $C_{o} \times T$ nodes whose topology is defined by its vertex and an adjacency matrix $A'_{C_{o}}$. The adjacency matrix represents the connectivity or relation between the different nodes (objects) and its weights represent how strong their relationship is at different time steps. Fig.~\ref{fig_gcn} represents an overview of this module. 

\subsubsection{\textbf{Graph reasoning}}
The graph reasoning aims to do cross-class reasoning on the previously constructed graph. The objects relations are video dependent, and so multiple GCN blocks are stacked to learn multiple levels of semantics. Moreover, the adjacency matrix is also parameterized so that it can be learned and optimized with the pipeline during the training phase. Moreover, it can learn to adapt to the data itself. We also make use of self-attention mechanisms. Consequently, our adjacency matrix learns better to differentiate class relations owing to different videos. Fig.~\ref{fig_gcn} represents a block of the graph convolution reasoning.

\par
As the object relations are complex, it is hard to predefine the inter-object relations for each video. Therefore, by leveraging the self-attention mechanism~\cite{vaswani2017attention,agcn}, our graph adjacency matrix is learnable and can vary with the videos.  
In practice, the adjacency matrix $A_{C_{o}}$ is initialized with a fully connected matrix. Finally, the full topology of our graph is $A_{C_o}$ $\in$ \hspace{0.1 cm} $\mathbb{R}^{C_o \times C_o}$  and  the vertexes  representation $G_{in}$ $\in$ \hspace{0.1 cm} $\mathbb{R}^{D_2 \times T \times C_o}$. First, we embed the input $G_{in}$ using bottleneck convolutional layer (i.e. 1 × 1), then the output feature maps are rearranged into  $\mathbb{R}^{D_2 \times T \times C_o}$ and  $\mathbb{R}^{C_o \times D_2 \times T}$ 
followed by a matrix multiplication. The value of the resultant matrix is then normalized by a softmax activation. Now, the superimposed adjacency matrix $A'_{C_{o}}$
can be formulated as:
\begin{equation} \label{eq:4}
A'_{C_{o}} = A_{C_{o}} + softmax(W_{1}^{T}G_{in}^{T}W_{2}G_{in})
\end{equation}
Where $W_{1}$ and $W_{2}$ are learnable weights of the bottleneck convolutions, and $G_{in}$ being $F'$  the stacked class representations in section B. $G_{out}$, the output of the graph layer is passed to the next graph layer and follows the same equations. In this work, we use 5 blocks of graph convolutions. As for the $A'_{C_{o}}$, each value represents an edge between two nodes (objects). We learn a graph that is shared across different time-steps but depends on each layer and for each video, as we said earlier we learn different semantics at each level.
\newline
After bottleneck convolutions, we do the graph convolution operation with the formulation in \cite{kipf2016semi}:
\begin{equation} \label{eq:5}
G_{out} = A'_{C_{o}}G_{in}W_{3}
\end{equation}
$W_3$ is a learnable parameter where $W_3$ $\in$ \hspace{0.1 cm} $\mathbb{R}^{D_2 \times D_2}$. The equation \ref{eq:5} represents the message passing and node feature updating, and finally $G_{out}$ is rearranged to $\mathbb{R}^{D_2 \times T \times C_o}$.\\
From equation \ref{eq:5}, we can understand how graph convolutions work. The graph convolutional layer represents each node as an aggregate of its neighborhood, hence each node gathers information from its neighborhood and adapts itself accordingly. In other words, at each graph block, each object collects information about other objects and finally finds to which it is most correlated, and thus whether there is an interaction or not. That is why we judge that the use of graphs is a promising idea in this domain.
\subsubsection{\textbf{TCN}} 
\begin{figure}[!t]
\includegraphics[width=\linewidth]{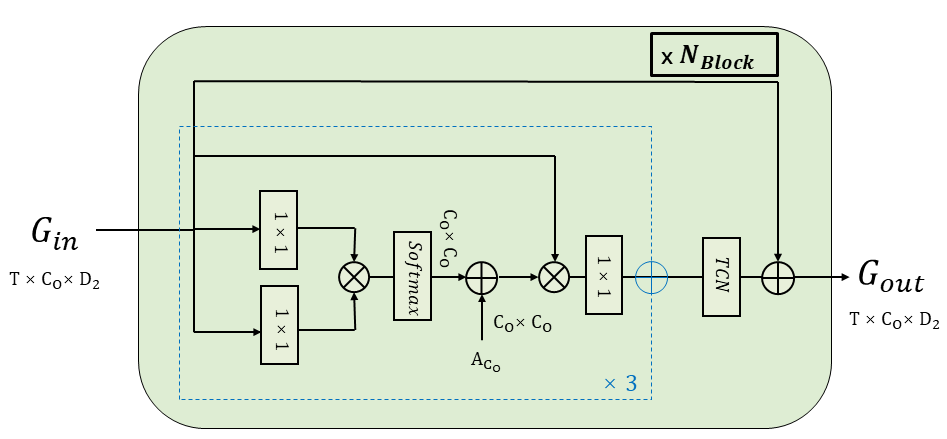}
\caption{Overview of one layer of the Object Relation Reasoning module, using a graph architecture~\cite{agcn}. As we can see, the input is a graph representation between different classes and the output is an updated representation of the graph. The $\times$ $N_{block}$ stands for the number of blocks used in total, while the $\times$ 3 at the bottom in blue stands for the number of used multi-head attentions.}
\label{fig_gcn}
\end{figure}

stands for Temporal Convolution Network. The graph reasoning is capable of extracting the relation between objects. 
However, in our study, we aim at modelling the spatio-temporal interaction in a large time span. To do so, we add a 1D convolution layer on top of the previous output of \textit{the graph reasoning} (i.e., $G_{out}$). As shown in Fig.~\ref{fig_gcn}, each \textit{Object Relation Reasoning Module} contains a \textit{TCN}. This 1D-convolution layer is used to aggregate the information across time. While stacking multiple object relation reasoning blocks, each block is used to model the object relation in a specific temporal scale. 
Finally, the output of the \textit{Object Relation Reasoning Module} is:
\begin{equation} \label{eq:6}
G_{out} = Conv1D(G_{out}) +G_{in}
\end{equation}
As mentioned earlier, the output of each block $G_{out}$ is the input $G_{in}$ to the next block.

\subsection{Predictions}

\begin{table*}[]
\begin{center}
\caption{Ablation study on different settings. This evaluation is on EPIC-KITCHEN dataset. Temporal nodes means using the final output of X3D of size $T\times2048$ to create nodes, while spatio-temporal nodes means using a mid layer of size $T\times7\times7\times432$ with more spatial information. Finally ADJ-matrix stands for using the adjacency matrix for predicting the verbs instead of using only nodes for nouns and verbs.}
\begin{tabular}{|l|cc|cc|cc|}
\hline
                                                  & \multicolumn{2}{c|}{verbs}       & \multicolumn{2}{c|}{nouns}       & \multicolumn{2}{c|}{actions}     \\ \hline
                                                  & \multicolumn{1}{c|}{top1} & top5 & \multicolumn{1}{c|}{top1} & top5 & \multicolumn{1}{c|}{top1} & top5 \\ \hline
\multicolumn{1}{|c|}{X3D}                         & \multicolumn{1}{c|}{46.5} & 79.8 & \multicolumn{1}{c|}{34.3} & 65.3 & \multicolumn{1}{c|}{21.0} & 38.7 \\ \hline
\multicolumn{1}{|c|}{THORN/temporal nodes} &
\multicolumn{1}{c|}{55.8} & 82.86 & \multicolumn{1}{c|}{39.9} & 66.37 & \multicolumn{1}{c|}{26.8} & 44.0 \\ \hline

THORN/temporal nodes + ADJ-matrix          & 
\multicolumn{1}{c|}{60.3} & 86.0 & \multicolumn{1}{c|}{41.1} & 66.9 & \multicolumn{1}{c|}{30.1} & 47.3 \\ \hline

THORN/spatio-temporal nodes + ADJ-matrix   &
\multicolumn{1}{c|}{61.0} & 85.9 & \multicolumn{1}{c|}{42.9} & 67.9  & \multicolumn{1}{c|}{30.5} & 47.5 \\ \hline
\end{tabular}
\label{table:abal}
\end{center}
\end{table*}

Predictions are based on the learned nodes and adjacency matrix.
However, since in our case our actions are composed of verbs and nouns, we show that using the adjacency matrix for predicting the verb and the object feature representation for noun prediction is more effective. This makes sense since the adjacency carries more information about how different objects interact with each others, while the nodes carry a refined object representations, after been processed through the different graph convolutions blocks. Our final layers are two fully-connected layers one projecting  $G_{out}$ from  $\mathbb{R}^{D_2 \times C_o}$ to  $\mathbb{R}^{1\times C_o}$, and the other fully-connected layer projecting $A'_{C_o}$ from $\mathbb{R}^{C_o \times C_o}$ into  $\mathbb{R}^{1 \times C_v}$, where $C_o$ and $C_v$ stand for the number of object classes and verb classes respectively.
\newline
Since we have 3 outputs, our loss is a sum of three losses and can be formulated as :
\begin{equation} \label{eq:7}
\mathcal{L} = \mathcal{L}_{verbs} + \mathcal{L}_{nouns} + \mathcal{L}_{clip-objects} 
\end{equation}
Where $\mathcal{L}_{verbs}$ and $\mathcal{L}_{nouns}$ are the negative log-likelihood losses (since each action is composed of one verb and one noun). As described earlier, the $\mathcal{L}_{clip-objects}$ is the loss to ensure the semantic of the object representation. 

\section{Experiments}
\noindent\textbf{Dataset}.
We have evaluated our model on two of the largest and challenging datasets for first-view  and human-object interaction action recognition. \textbf{Epic-Kitchen55 }\cite{damen2018scaling} contains 55 hours of recording of 32 different kitchens in 4 cities. This dataset has a total of 125 verbs and 352 nouns. 
\textbf{EGTEA Gaze+}~\cite{gaze} contains 28 hours of cooking activities from 86 unique sessions of 32 subjects, with over 10k video clips of 106 fine-grained egocentric activities.
In both datasets, each action is a combination of a verb and a noun. Actions are relevant to different steps of preparing food (e.g. \textit{cleaning the kitchen}, \textit{cutting vegetables}, \textit{preparing table}).
\newline
\textbf{Implementation}.
We implement our method using X3D as the visual encoder where $D_1$ = 432, $H'$= $W'$= 7 and $D_2$ is 128. We input a clip of 16 RGB frames for Epic-Kitchen and 25 frames for EGTEA Gaze+. We use a dropout probability of 0.3. For the \textit{object relation reasoning} module, $N_{Block}$ is 5 blocks. 



For the temporal convolution network, we run our model with different values of the kernel size. As there was no impact on the results, we kept a kernel size of 9.
In training phase, we utilized Adam \cite{adam} to optimize the model with an initial learning rate of 0.00005. We scaled the learning rate by a factor of 0.1 with the patience of 5 epochs. The network was trained on a 4-GPU machine for 30 epochs. We evaluated our model using top1 and top5 accuracy on verbs and nouns for Epic-Kitchen, while for EGTEA Gaze+ we evaluated directly on actions using top 1 accuracy.

\subsection{Ablation Study}
In this section, we validate our model design for the modules in the THORN. The evaluation is conducted on the EPIC-Kitchen dataset. We propose different settings, and see how each setting can improve the performance. 
In table ~\ref{table:abal}, we can notice different results:

\par
Firstly, we compare our baseline model X3D with THORN. 
Note that, in THORN, the graph nodes can be constructed either using the output of the last layer of X3D (temporal nodes) or using its intermediate layer (spatio-temporal nodes). Here, we first compared X3D with THORN (temporal nodes), i.e., we construct the nodes by the features in shape $T\times2048$. 
In this setting, nodes would serve to predict both verbs and nouns. In this scenario, we improve nouns prediction by \textbf{+5.6\%}, while, the verbs accuracy increased by \textbf{+9.3\%} . Proving the importance of the cross-object reasoning, compared to only capturing visual information from 3D-CNNs.\\
Secondly, we study the importance of the adjacency matrix for predicting the verbs. To do so, we use the adjacency matrix (ADJ-matrix) to predict verbs, while keeping the nodes to predict the nouns.
In this setting, the verb prediction improves by \textbf{+4.5\%} compared to the previous setting and by \textbf{+13.8\%} to the baseline X3D. This is because the adjacency matrix captures the object interaction, hence, it is more suitable for verb prediction. \\
Thirdly, we study the effect of changing the temporal nodes with the spatio-temporal nodes. 
Spatio-temporal nodes are the nodes constructed by the middle layer of X3D which contains the spatial information ($T\times7\times7\times432$).   
With spatio-temporal nodes, THORN improves   \textbf{+1.8\%} on nouns. This is because, with spatial dimensions, the ORF can better capture the object relative locations and the size of the object, then embed them in the node representation. As a result, the noun accuracy improves. This setting also brings  \textbf{+0.7\%} improvement on verbs.
\\
Our overall architecture obtains \textbf{+13.8\%} more accuracy on verbs and \textbf{+8.6\%} on nouns w.r.t. vanilla X3D. This reflects the importance of our proposed modules in THORN and how an object-centric method can improve results on human-object interaction actions.
%

\begin{table}[]
\centering
\caption{Ablation study on fusing the scores of THORN with the scores from the object detector (Faster RCNN). 
This evaluation is on EPIC-KITCHEN dataset.
Fusing both scores brings significant improvement on top-1 accuracy. For the object detector, we use an average pooling on all the video clip frames object detection scores and add a thresh-hold of 0.3 }
\label{faster}
\begin{tabular}{|c|c|l|}
\hline
Faster-RCNN scores          & THORN                  & \multicolumn{1}{c|}{Nouns} \\ \hline

\checkmark              & $\times$                     & 31.5                       \\ \hline
$\times$                      & \checkmark             & 32.8                       \\ \hline
\checkmark               & \checkmark             & 42.9                       \\ \hline
\end{tabular}
\label{table:nouns}
\end{table}
We then study the components for predicting the nouns in our model. In table~\ref{table:nouns}, we show that fusing scores of object detection and the scores obtained by the THORN nodes representation works better than using only one of them. 
We also find that predictions using only our model are better than the object detector itself. This shows that our model can refine the objects represented by the other objects (nodes) using our graph-based module. 

\subsection{Comparison with the State-of-the-Art}
We then compare our proposed method with the state-of-the-art methods on EPIC-Kitchen and EGTEA Gaze+ in table~\ref{table:comp_sota} and~\ref{table:egtea+}.

In Table~\ref{table:comp_sota}, we compare our results with the state-of-the-art methods. Among these methods, Long Features Bank (LFB) \cite{LFB} proposes to use global as well as local features for action recognition. To do so, they extract features on both clip and video levels, and combine them to have a better understanding of the scene. Nevertheless, this method still lacks accuracy for the objects. Moreover, LFB is a two step method which trains separately an object and verb recognizer modules. For our THORN, we train a single model for predicting both entities. As a result, we have a \textbf{+8.5\%} improvement on top 1 nouns and a \textbf{+4.9\%} w.r.t. LFB on action recognition. 
%

Our method achieves the overall best performance. We claim that AssembleNet++ utilizes additional modality such as optical flow in both training and inference time. Even though, we still have the lead in top 1 accuracy for the verbs, nouns and actions, which proves again that having an object-centric and specific reasoning on object interactions is a key solution for having a better action recognition on HOI datasets. Finally, our results prove that using only RGB with an object-centric model achieves better or similar results compared to methods relying on heavy multi-modality reasoning.


In table~\ref{table:egtea+}, we compare our method with the state-of-the-art on EGTEA Gaze+ dataset. We have the best accuracy w.r.t. the others methods, which shows the generalization and robustness of our model on actions of HOI.

To sum up, compared to other methods, ours is lightly weighted as we use X3D, while other methods rely on heavy 3D-CNNs such as I3D. THORN is trained jointly on nouns and verbs as opposed to other methods such as LFB~\cite{LFB}, and we only need RGB frames and object classes per-frame.\\
\begin{table}[]
\begin{center}
\caption{Comparing THORN model with other state-of-the-art methods on the validation set. Even though some of these comparisons are not fair since these models are using multi-modalities, we still hold the best accuracy on actions and nouns, which shows the strength of our model}
\scalebox{0.75}{
\begin{tabular}{|c|c|c|c|c|c|c|c|}
\hline
Model                                              & Obj                   & RGB       & \multicolumn{1}{l|}{Flow} & \multicolumn{1}{l|}{Audio} & \multicolumn{1}{l|}{Verbs} & \multicolumn{1}{l|}{Nouns} & \multicolumn{1}{l|}{Actions} \\ \hline
                                                   & \multicolumn{1}{l|}{} &           & \multicolumn{1}{l|}{}     & \multicolumn{1}{l|}{}      & \multicolumn{1}{c|}{top1}  & \multicolumn{1}{c|}{top1}  & \multicolumn{1}{c|}{top1}    \\ \hline
Baradel\cite{baradel2018object}                                            & $\times$                     & \checkmark & $\times$                         & \checkmark                  & 40.9                       & -                          & -                            \\ \hline
3D-CNN                                             & $\times$                     & \checkmark & $\times$                         & $\times$                          & 49.8                       & 26.1                       & 19.0                         \\ \hline
STO\cite{LFB}                                                & \checkmark             & \checkmark & $\times$                         & $\times$                          & 51.0                       & 26.6                       & 19.5                         \\ \hline
LFB\cite{LFB}                                            & \checkmark             & \checkmark & $\times$                         & $\times$                          & 52.6                       & 31.5                       & 22.8                         \\ \hline
\multicolumn{1}{|l|}{AssembleNET++ ODF+SDF\cite{Anet}} & \checkmark             & \checkmark & \checkmark                 & $\times$                          & \textbf{60.0}              & 37.1                       & 25.2                         \\ \hline
\textbf{THORN}                                      & \checkmark             & \checkmark & $\times$                         & $\times$                          & \textbf{61.0}                      & \textbf{42.9}              & \textbf{30.5}                \\ \hline
\end{tabular}
}
\label{table:comp_sota}
\end{center}

\end{table}

\begin{table}[]
\begin{center}
\caption{Comparing THORN model with other state-of-the-art methods on  EGTEA Gaze+ split1. We  hold the best accuracy on actions}
\scalebox{0.7}{
\begin{tabular}{c|c|c|c|c|c|c|c|}
\cline{2-8}
                             & Two-stream & I3D \cite{3D-cnn-2}  & TSN \cite{tsn}  & ego-rnn \cite{ego} & LSTA \cite{lsta} & SAP \cite{sap}  & \textbf{THORN} \\ \hline
\multicolumn{1}{|c|}{ACC \%} & 43.8       & 54.2 & 58.0 & 62.1    & 62.0 & 64.1 & \textbf{67.5}  \\ \hline
\end{tabular}
}
\label{table:egtea+}
\end{center}
\end{table}

\subsection{Qualitative Study}
In this section, we conduct a qualitative study of THORN. 
\begin{figure}[!t]
\includegraphics[width=\linewidth,height=3cm]{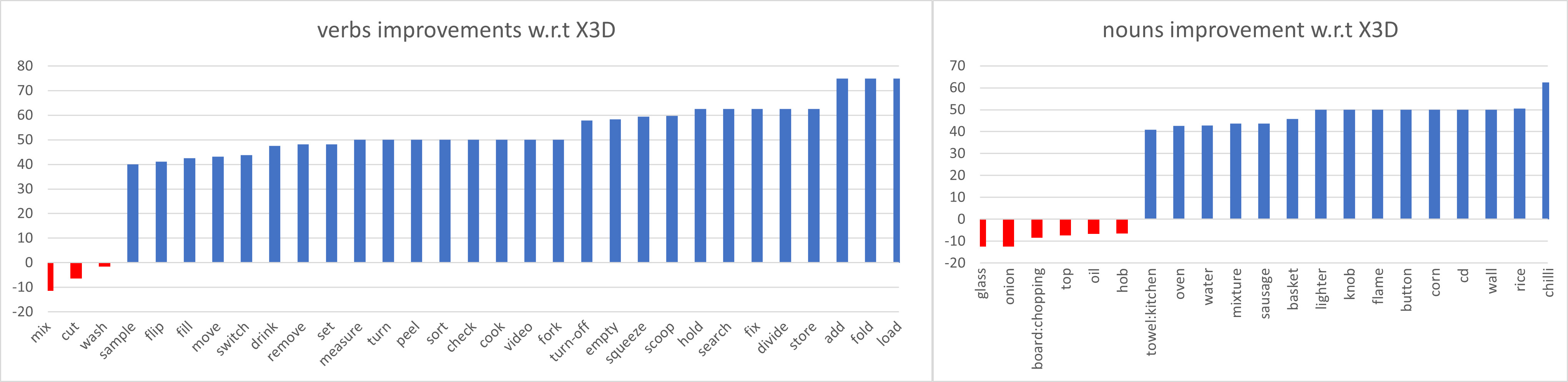}
\caption{Accuracy improvement on nouns (right) and verbs (left)  w.r.t X3D.}
\label{fig_imp}
\end{figure}
%
In Fig.~\ref{fig_imp}, we show the impact on some classes after adding our proposed module w.r.t. vanilla X3D. In EPIC-Kitchen, we significantly improve accuracy on 28 verb classes. 
Only the accuracy of 3 out of 125 verbs decreases, while the decrease is negligible. 
This improvement on verbs shows that understanding the inter-relation of different objects is important for HOI.

For noun recognition, it is interesting to find that THORN can now predict some classes such as \textit{water} and \textit{wall}. These classes are barely detected with the object detector. This is a result of the reasoning process on cross object classes, which refines the nodes and can finally predict overlooked object classes. 
\par
\subsection{Qualitative Study on Learned Adjacency Matrix}
\begin{figure}[h!]
\includegraphics[width=\linewidth, height=4cm]{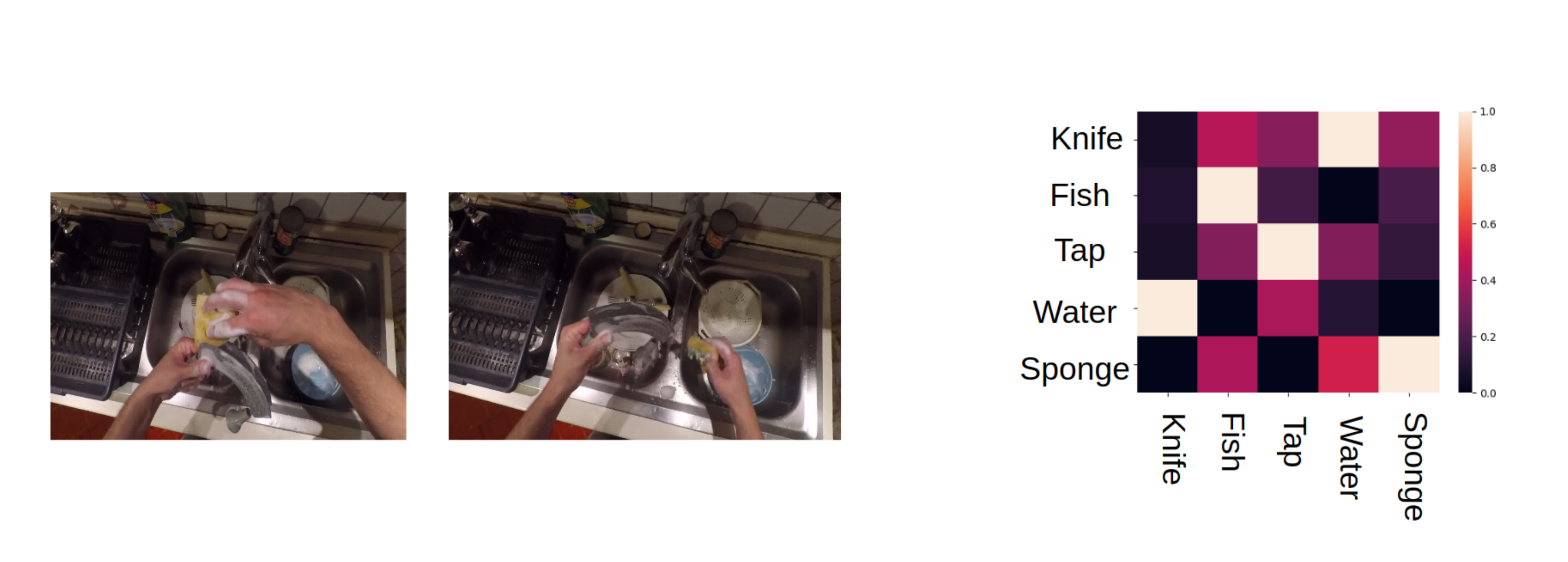}
\caption{Example of the learned adjacency matrix of the action  from Epic-Kitchen55 dataset. We notice a strong correlation between the classes ~\textit{knife} and \textit{water} for the action  \textit{wash knife}. Thus, we are able to collect high inter-class relation to recognize the right verb and its relevant objects. Moreover, the irrelevant classes such as \textit{fish} are not activated, showing robustness of the learned attention.}
\label{fig_attention1}
\end{figure}

In this section, we provide more insight of our THORN model. We show the strength of using the adjacency matrix and the attention mechanism.

In Fig. \ref{fig_attention1} we show an example of the learned adjacency matrix for the action \textit{wash knife}. In this figure, we find that there is a high correlation between the classes \textit{knife} and \textit{water} in both directions. Whereas the classes \textit{tap, fish} and \textit{sponge} are only correlated to themselves since they are not directly relevant to the objective action class \textit{wash knife}. This example shows the effectiveness of THORN to capture the inter-object relations in the clipped HOI videos. 
\\
\subsection{Qualitative Study on the  Object Representation Filter}
\begin{figure}[h!]
\includegraphics[width=\linewidth]{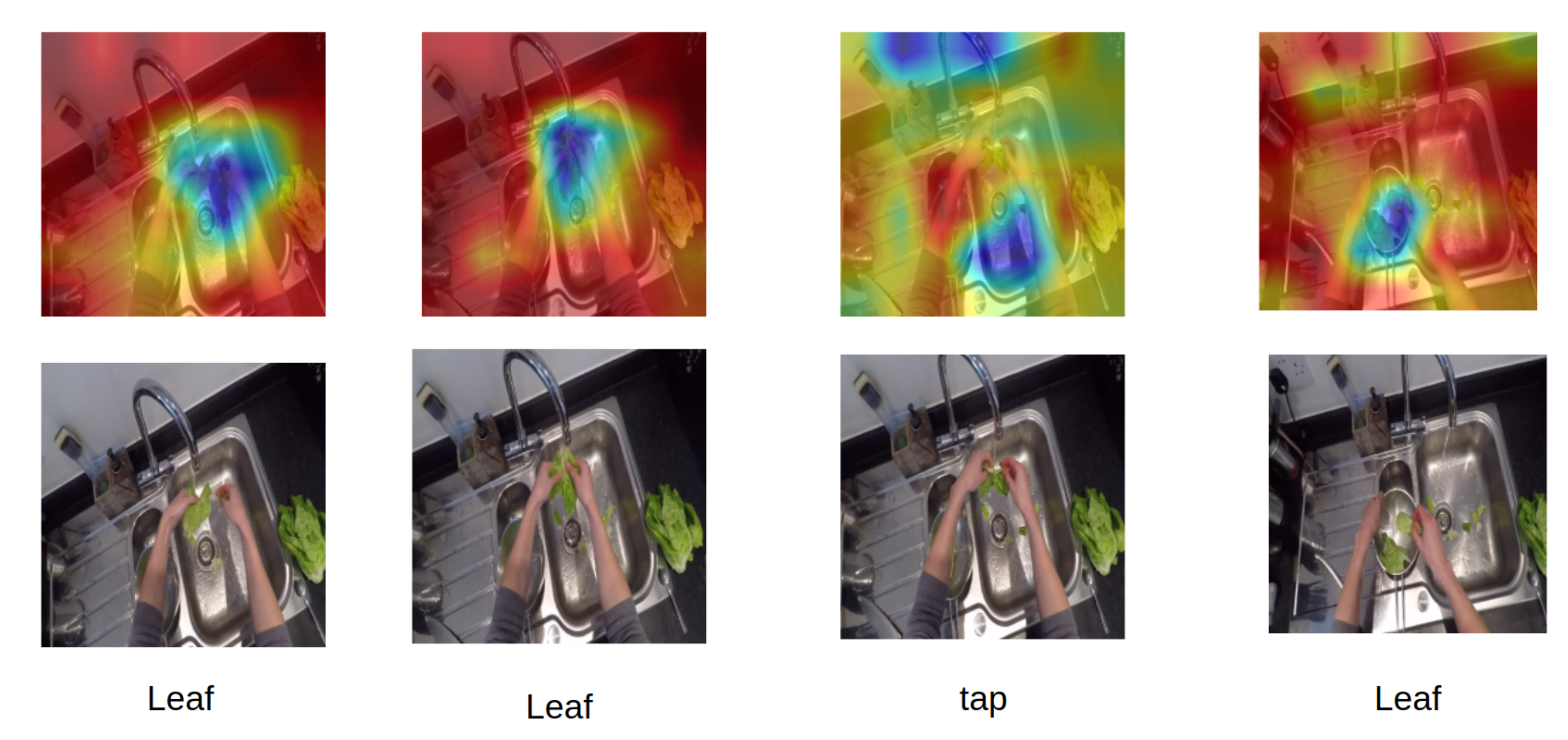}
\caption{Example of action \textit{washing leaf}. the highest activated classes were leaf and tap and when inferring the class activation map we can see that most activated pixels are around the objects of interest. Hence, the features extracted are more significant which makes it easier to predict the right action.}
\label{fig_0}
\end{figure}

\begin{figure}[h!]
\includegraphics[width=\linewidth]{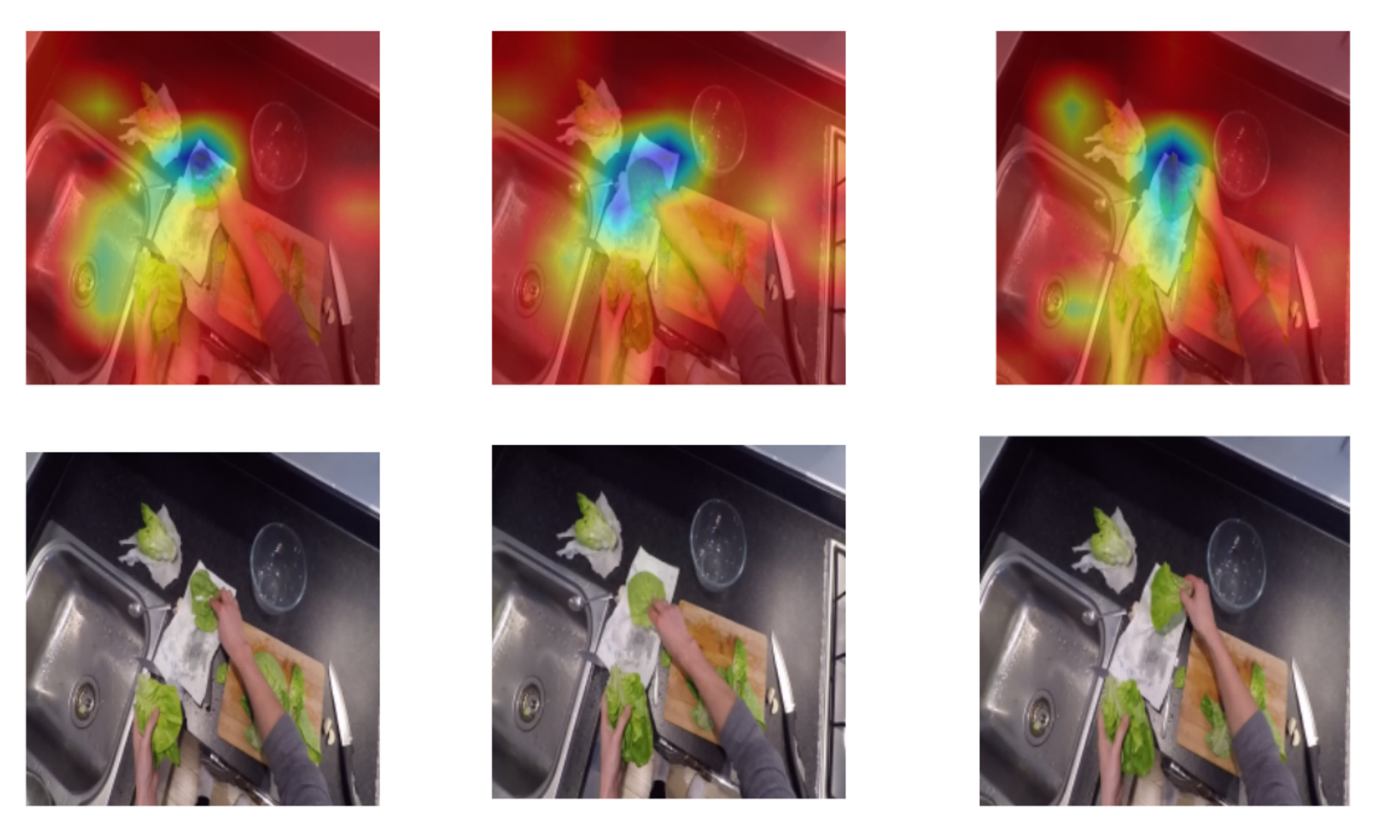}
\caption{Example of action \textit{put leaf}. In this example the most activated object was leaf and its activation map shows that the focused-on pixels actually belongs the leaf, proving the strength and robustness of our approach.}
\label{fig_1}
\end{figure}

\begin{figure}[h!]
\includegraphics[width=\linewidth]{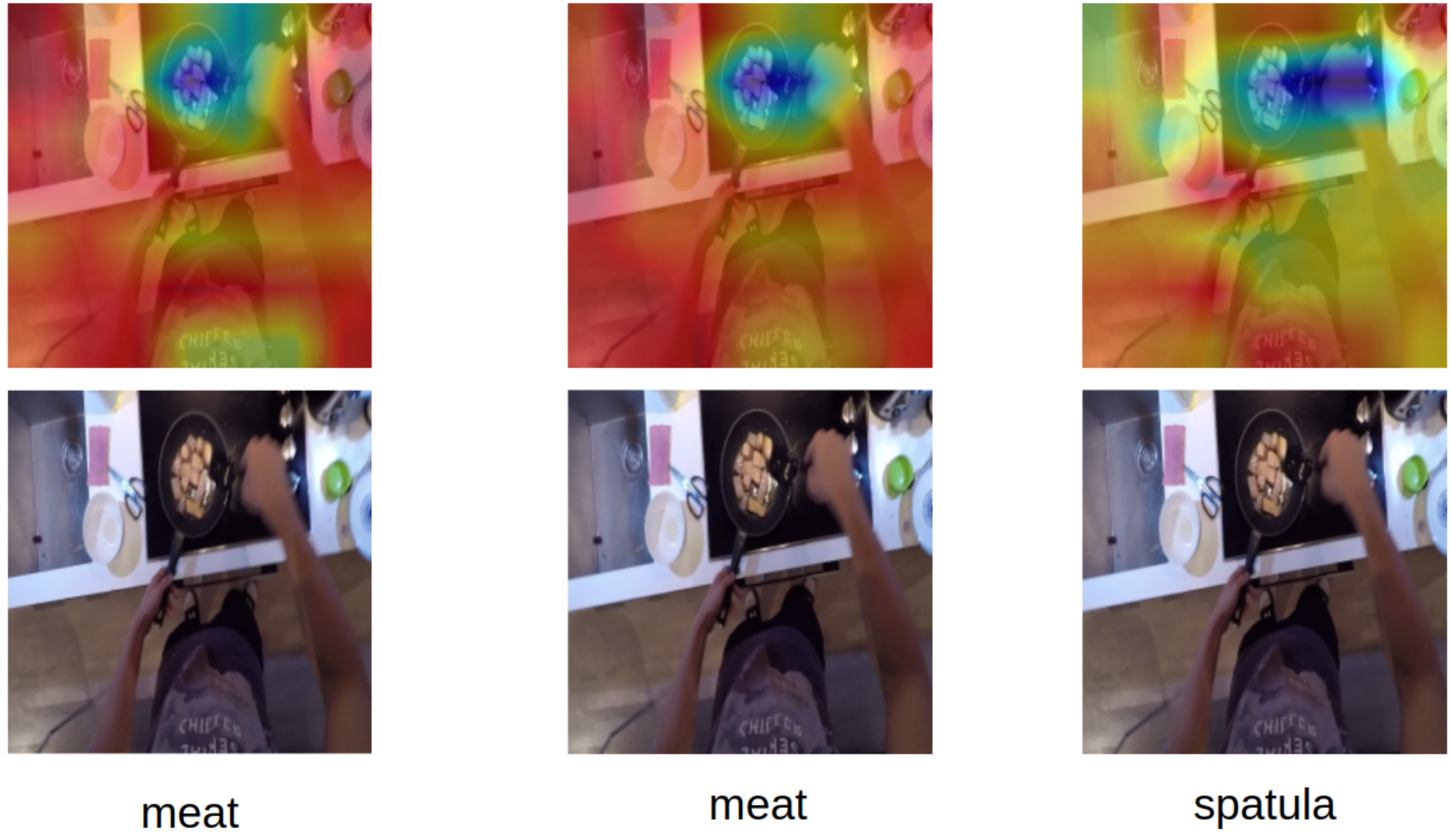}
\caption{The action in this figure is \textit{mix meat}, and looking at the figure we notice that the highlighted pixels are the ones corresponding to the spatula and the meat. Therefore, it is easier to predict the right action. }
\label{fig_2}
\end{figure}

The object representation filter is one of the main parts of our architecture as it allows extraction of a good representation for different objects related to the action. To make sure our filtering work, we extract the activation maps for the different object and see what do they highlight in the scene.\\
Figures ~\ref{fig_0},~\ref{fig_1},~\ref{fig_2}, represent different actions with their Class Activation Map (CAM). The example in Fig.~\ref{fig_0} represents the action \textit{wash leaf}, when looking at the output of the object representation filter the highest activation where on the classes \textit{leaf} and \textit{tap}. As specified in the main paper, we want to learn features specific to each class. The CAM of tap and leaf in this example clearly shows that only the pixels relevant to the object were highlighted, hence, the feature in the nodes are more representative of the objects of interest.
\newline
Moreover, this result shows that our work does similar work to unsupervised object segmentation. Hence, unlike other methods that rely on pre-trained object detectors and tracking methods to extract object and then use ROI-Align to extract objects features, our method is capable of yielding the same result in a unsupervised manner and in a more simplified way. Besides that, our THORN model learns to only focus on objects of interest.
\newpage

\section{Conclusion}
First-view action recognition relies on capturing the visual relationships between different objects and the human. In this work, we propose an object-centric model, which projects the standard CNN features into object class-specific features. After that, we compute the inter-object relations in graph reasoning, where each node corresponds to an object class and each edge represents the relation between two different objects. We evaluate our model on two large and challenging datasets. THORN achieves state-of-the-art performance on both datasets, which shows the effectiveness and robustness of our method. 
As our method relies on object detection precision,
our future work aims at developing an architecture that can combine object detection and action recognition tasks. We also want to extend our model for first-view action detection for untrimmed video.







\newpage
\bibliographystyle{IEEEtran}
\bibliography{IEEEabrv}
\newpage

%

\end{document}